\documentclass[3p,times,preprint]{elsarticle}
\usepackage{ecrc}
\volume{00}
\firstpage{1}
\journalname{XXX}
\runauth{M. Zocholl et al.}
\jid{procs}

\usepackage{amssymb}
\usepackage[figuresright]{rotating}
\usepackage{amsmath,amssymb,amsfonts}
\usepackage[table,usernames,dvipsnames]{xcolor}
\usepackage{latexsym}
\usepackage{epsf,epsfig,float,graphicx}

\usepackage{dirtree}

\definecolor{redish}{rgb}{0.75, 0.3, 0.40}
\begin{document}
\begin{frontmatter}

\dochead{Pre-print and extended version of the poster paper presented at the 14th International Conference on Semantic Systems}
\title{Ontology-based Design of Experiments on Big Data Solutions}
\author[a]{Maximilian Zocholl\corref{cor1}} 
\author[a]{Elena Camossi}
\author[a]{Anne-Laure Jousselme}
\author[b]{Cyril Ray}
\address[a]{NATO STO Centre for Maritime Research and Experimentation (NATO STO-CMRE), 
La Spezia, Italy}
\address[b]{Institut de Recherche de l'\'Ecole Navale (IRENav), Brest, France}

\begin{abstract}

Big data solutions are designed to cope with data of huge Volume and wide Variety, that need to be ingested at high Velocity and have potential Veracity issues, challenging characteristics that are usually referred to as the ``4Vs of Big Data".  In order to evaluate possibly complex big data solutions, stress tests require to assess a large number of combinations of sub-components jointly with the possible big data variations. A formalization of the Design of Experiments (DoE) on big data solutions is aimed at ensuring the reproducibility of the experiments, facilitating their partitioning in sub-experiments and guaranteeing the consistency of their outcomes in a global assessment. In this paper, an ontology-based approach is proposed to support the evaluation  of a big data system in two ways. Firstly, the approach formalizes a decomposition and recombination of the big data solution, allowing for the aggregation of component evaluation results at inter-component level. Secondly, existing work on DoE is translated into an ontology for supporting the selection of experiments. The proposed ontology-based approach offers the possibility to combine knowledge from the evaluation domain and the application domain.
It exploits domain and inter-domain specific restrictions on the factor combinations in order to reduce the number of experiments. Contrary to existing approaches, the proposed use of ontologies is not limited to the assertional description and exploitation of past experiments but offers richer terminological descriptions for the development of a DoE from scratch. As an application example, a maritime big data solution to the problem of detecting and predicting vessel suspicious behaviour through mobility analysis is selected. The article is concluded with a sketch of future works.
\end{abstract}

\cortext[cor1]{Corresponding author. Tel.: +39-0187-527-234; fax:+39-0187-527-1.}
\end{frontmatter}
\email{maximilian.zocholl@cmre.nato.int}

%=PAPER============================================================%
\section{Introduction} \label{sec:intro}

In this article, we investigate the use of knowledge based methods for selecting a suitable Design of Experiments (DoE) aiming at evaluating the performance of big data systems including human factor interactions.

Formalising scientific and engineering experiments makes them reproducible, easier to share and search. It lets their results get connected to experimental outputs of future studies and accelerates their interpretation. In order to make these principles easily applicable, different possibilities for their formalisation are subject to ongoing efforts. The contribution of this article is the initialisation of an enrichment of priorly proposed concept descriptions, up to the point where a DoE can be proposed by a non-expert. The approach is not claiming comprehensiveness but advocates a stronger use of semantic web technologies for making DoE specific knowledge accessible to other domain descriptions.

Design of Experiments is a collective of principles, statistical approaches and models for planning and performing experiments as well as analysing their results \cite{Fisher.1937}, \cite{Montgomery.2013}. Typically, the subject of experiment is reduced to a system with input and output variables. Some or all controllable input variables, the so called factors, are varied according to an experimental plan which specifies the values of the variables, also called factor levels. After feeding the system with a set of input variables the output is observed. As the output depends on the system behaviour and both controllable and possibly uncontrollable input variables, the goal of DoE is the quantification of the functional relation between the input and output of the system. For the functional description of different subjects of experiment and different hypotheses to test, a large body of knowledge exists either for extracting the maximum amount of information from a given number of experiments or for minimising the number of experiments with respect to a required level of confidence in information.

Two main challenges arise during the assessment of a big data system with a Design of Experiments. Firstly, big data variations translate seamlessly into a large number of factors with a multiplicity of possible factor levels. Secondly, the different components of the big data system implement either deterministic or non-deterministic processes and yield different output types, e.g. continuous or multinomial. Thus, for a thorough assessment the system needs to be unfolded into its components. Again, this increases the number of necessary experiments but additionally and more importantly introduces the necessity of using completely different types of DoE. 

For addressing the different types of DoE, an ontology based method is proposed to select suitable DoEs, based on the properties of the components. For the subsequent reduction of the number of experiments knowledge of the  
application domain and knowledge of the DoE domain are combined. After the performance of the experiments, their results are interpreted %AL - on --> at
at the component level by the same ontology based method which allows alternative interpretations and proposes the performance of additional experiments depending on the experimental results. %AL - On --> at
At the system level, the results of the component level experiments are recombined and % AL - enable --> is enabled (at the end of the sentence)
the formulation of hypotheses for the system behavior is enabled.

To evaluate %AL - the approach we propose, 
our approach, we consider a big data system developed to support Maritime Situation Awareness (MSA). 
For maritime %AL - + surveillance
surveillance operators situational awareness is necessary to assert a remotely happening situation correctly. As the actual situation at sea is only partially observable, the operators have to base their estimates on different data sources with different degrees of liabilities. Typical tasks for operators are the prevention of collisions, the identification of vessels in distress, the verification and observation of the respect of protected areas and fishing pressure, as well as %AL - detecting --> the detection of 
illicit activities and human trafficking at sea.

Each of these situations corresponds to a specific meaning. During operations an actual situation needs to be verified by the operator according to all possible interpretations of the available but incomplete and often incorrect data. While investigating one situation, the operator is less responsive to other simultaneously happening situations. Thus, Maritime Situational Indicators (MSI) shall annotate data in order to allow for a faster awareness creation and to %AL - allow for --> ease
ease a prioritisation of the actual situations. 
%\ec{discuss role of MSI in the proposed design, here or in the system decomposition section}

%This article investigates knowledge based methods for assessing big data systems for maritime situational awareness. Two main challenges arise during the assessment of big data system with a Design of Experiments (DoE). Firstly, big data variations translate seamlessly into a large number of factors with a multiplicity of possible factor levels. Secondly, the different components of the big data system implement either deterministic or non-deterministic processes and yield different output types, e.g. continuous or multinomial. Thus, for a thorough assessment the system needs to be unfolded into its components. Again, this increases the number of necessary experiments but additionally and more importantly introduces the necessity of using completely different types of DoE. 

%For addressing the different types of DoE, an ontology based method is proposed to select suitable DoEs, based on the properties of the components. For the subsequent reduction of the number of experiments knowledge of the maritime domain and knowledge of the DoE domain are combined. After the performance of the experiments, their results are interpreted on the component level by the same ontology based method which allows alternative interpretations and proposes the performance of additional experiments depending on the experimental results. On the system level, the results of the component level experiments are recombined and enable the formulation of hypothesis for the system behavior.

After the overview of existing contributions to the domain of ontologies in DoE and ontologies for situational awareness in \ref{sec:rw}%AL. T
, the main part of the article describes the system decomposition in \ref{sec:SD} and the ontology based method for DoE in \ref{sec:OE}.

%~~~~~~~~~~~~~~~~~~~~~~~~~~~~~~~~~~~~~~~~~~~~~~~~~~~~~~~~~~~~~~~~~~%
\section{Related Work} \label{sec:rw}

Important contributions come from the field of design of experiments and maritime situational awareness, situation assessment or domain awareness. Both domains show an increasing adherence to languages with well-understood semantics like first-order-logic and description logic, as well as newly developed extensions of those.

\subsection{Semantic representations of DoE}\label{sec:rw-doe}
In the domain of DoE existing formalisations use non-probabilistic languages, at most. As the described evaluation process focuses on big data solutions, the listed literature mainly originates from the domains of computer experiments and human machine interface experiments.

In the domain of computer experiments Do et al. provide an overview of empirical techniques for software testing identifying two approaches \cite{Do.2005}. Firstly, controlled experiments rely on the precise variation of given variables. Secondly, and complementarily, case studies 
follow possible scenarios of usage. Both approaches aim for the replicability of the performed experiments, the possibility to aggregate their results beyond the anticipated scope of design of experiments and by these means to validate the significance of their results in form of models. 
Precursors for enabling these benefits can be seen in the interpretability of the experimental results, e.g. by the documentation or standardisation, and in infrastructures allowing to share and connect artifacts \cite{Do.2005}, \cite{MLSchema}. In the data mining and machine learning domain recent efforts lead to the W3C ML Schema Community Group with the goal to support the development of a data exchange standard for experimental data by unifying existing, more specific schemata \cite{MLSchema}. More specifically, the group pools former efforts on DMOP \cite{Keet.2015}, Expose, the OpenML related Ontology of Vanschoren \cite{Vanschoren.2012}, as well as the contributions of Soldatova et al. in the form of EXPO and OntoDM \cite{Soldatova.2006}, \cite{Panov.2014}. For the formalisation of scientific experiments, especially \cite{Soldatova.2006} is of interest.

Another approach from Blondet et al. propose a DoE ontology in the context of the product development process. The aim is to describe executed DoE, thus to make them queryable during the design phase of later experiments \cite{Blondet.2018}. For the definition of a DoE three sampling methods, called 'types of DoE' are differentiated, namely marginal constraints like Latin Hypercube, factorial designs and low discrepancy or quasi random, e.g. Halton \cite{Blondet.2016}. Subsequently further information requirements are added, including the number of experiments, an initial model, one or more factors, one or more outputs, a surrogate model and some analysis methods \cite{Blondet.2018}. Also constraints like the limited amount of time or thresholds like accuracy of the predictions and surrogate model predictivity are supposed to be specified \cite{Blondet.2016} and extended by a maximum number of experiments \cite{Blondet.2018}. Nevertheless, the proposed rules and concept descriptions do not allow for the deduction of a subset of DoE types fulfilling the set of constraints. Instead, SPARQL queries are used to find existing DoE instances which are similar in the sense of generalisation. An assessment of the quality of the DoE instances is not 
performed, nor is it described which constraints exclude certain DoE.

All reviewed contributions aim at the description of experiments with respect to the proposed concepts. Complementary to this, the following paragraphs propose to expand the existing formalisations on knowledge from the domain of Design of Experiments that helps 
exclude designs which are not appropriate for a given set of constraints on the experiments. Especially, the knowledge is supposed to be captured in the T-Box of the Ontology, thus supports the Design of Experiments which start without prior knowledge of the domain or instances in the A-Box.

\subsection{Semantic representations of Maritime Situational Awareness}\label{subsec:rw-msa}
In the maritime domain contributions towards situational awareness exist from solely probabilistic approaches to purely possibilistic approaches, like \cite{Roy.2010} and \cite{VanDenBroek.2011}. Promising approaches combine the benefits of both worlds. One example is the work of \cite{Snidaro.2015} which uses Markov Logic Networks, i.e. weighted first-order-logic rules, for detecting maritime events. Also the work on Bayesian extensions of OWL \cite{Carvalho.2011} and \cite{Laskey.2011} is closely related to MSA as well. Related works can also be found in the avionic domain, e.g. describing data uncertainty and veracity for situation assessment with OWL \cite{Insaurralde.2017}.
%AL - Pour la version finale, j'aimerais bien revoir cette partie et rajouter quelques references, et en enlever certaines

%AL - General comments:
% - Synopses are not introduced
% - An exemplar list of MSIs could be provided
% 
\section{System decomposition and recombination}\label{sec:SD}

Typically, the quantification of a systems properties with designed experiments is done on the entire system taking into account the whole array of variations of the input variables. For reducing the DoE model complexity and the number of experiments, in the proposed approach both the input data and the system are first decomposed, then recombined to obtain an evaluation at inter-component level. Contrary to wide-spread systems engineering approaches and to systematic product testing and development methods, the proposed methodology aims for the initialisation of the evaluation subsequent to the component deployment but prior to the finalisation of the big data system. 
The assumption is that the system is developed according to a modular architecture, where system components are independently implemented then integrated in a second stage into the final big data solution. In this case it is not necessary to wait for the complete solution deployment to start the evaluation but the design of experiments leverages the testing of the modular components. In addition, the experiments on the complete solution may be customised depending on the performance of the single components. 

\subsection{Input description}
Starting from the system level, the research questions ask for the possible interference of performance of the system when confronted to typical dataset or application specific variations of any of the big data dimensions \cite{Kitchin.2016}. 

In the context of Maritime Situation Awareness, surveillance systems merge, process and analyse information of various type, collected from multiple sources of different nature. These include physical sensors, automated systems, as well as persons. To give some examples, Terrestrial-Automatic Identification System (T-AIS), High Frequency (HF)-Radars, Long Range Tracking and Identification (LRIT) systems, CTD (Conductivity, Temperature, ``Depth'') sensors, sea state models, underwater hydrophones, tracking software, human operators and analysts, are all sources of information that may be elaborated by surveillance system with the aim of increasing awareness. Real-time data streams from surveillance and environmental sensors are merged and put in context with databases, registries and intelligent reports, according to the mission and the specific operational task at hand. 
Because of the rapid growth in the use of AIS data transceivers on vessels and the surge in satellite received AIS data, data-driven maritime event detection \cite{Patroumpas.2015} based on AIS have been developed, as well as data-driven approaches in support to environmental protection and informed policy making \cite{Natale.2015}. Also, many commercial systems exist which use AIS as main source of information.  

All this information has different formats, and could be structured or semi-structured according to specification and standards, as in the case of AIS, or completely unstructured. 

Maritime surveillance systems present many aspects of interest for big data experiments, with the most important big data properties %AL +++
(and corresponding challenges) 
being volume, velocity, variety and veracity. Even considering systems based mainly on AIS, infrastructures able to to cope with high volume and velocity are necessary. On the other hand, due to the nature of AIS data, the veracity of the data can vary dramatically but can also be %AL 
better assessed 
%AL - better and better 
by using a variety of data sources from different data source types \cite{Snidaro.2015}. Each component of a maritime surveillance system is subject to the performance assessment of one or multiple dimensions of big data variation. 
%AL - For the final version, I would be happy to add some references

% Initial problem: way too many factors and at least 2 levels per factor.
% Candidates for Factors:
% - Input: AIS Data (dynamic data, Messages 1,2,3,5,18,19,24: with n different fields \cite{AIS.2014})
% - Input: dynamic data (AIS, synthetic aperture radar), contextual information (Port information, Nautical charts, navigation rules and guideline, regulated areas, vessel registers, blacklists, fisheries and vessel facts), sea and weather condictions, maritime features (black hole regions, maritime routes)
% - User-selected Parameters: Thresholds of MSI, combinations of MSI
% - Processing parameter: number of threads, cores, etc.

\subsection{System decomposition}

%In order to quantify the system performance component-wise %or to propose changes for performance improvements, 
%a %more 
%detailed characterisation of the subsystems is needed. Additionally, the 
%component-wise decomposition allows for the simultaneous validation of deployed subsystems while other subsystems are still under development. Thus, after measuring the performance of experiments on one subsystem, changes on this subsystem do not necessarily imply the expiration of the results on a higher level of component aggregation but result in a well-framed set of additional experiments.

%The system is decomposed into subsystems according to its %the 
%topology, % of the system, 
%in this case 
%for instance according to the interfaces to the publisher subscriber architecture %(e.g. CER) 
%and for a refined decomposition according to the functionalities. %(e.g. LED, SG) or (e.g. SI -> Weather integrator, RDFiser, link discovery)
%Each subsystem ingests an input, performs one or multiple functions %on the input 
%based on that 
%processing it and finally, issues an output. The output of one subsystem %becomes 
%is typically the input of another subsystem.

In order to evaluate the system performance, the system is %AL - breakdown 
broken-down along two, complementary, functional dimensions. 
%The system, which may be decomposed into subcomponents according to its architectural topology, may be evaluated component-wisely accordingly. At the same time, components and systems may be evaluated semantically-wisely, with the respect to the amount knowledge they contribute to increase awareness.  
On the one hand, it is divided into the software components it is made of, according to the designed software architecture. Such a decomposition enables the independent and simultaneous evaluation of deployed subsystems while other subsystems are still under development. 
The system is decomposed into subsystems according to its  
topology, for instance according to the interfaces to the publisher-subscriber architecture and for a refined decomposition according to the functionalities. 
Each subsystem ingests an input, performs one or multiple functions 
processing it and finally, issues an output. The output of one subsystem 
is typically the input of another subsystem.
%AL - Removed the line to remove the new paragraph
The independence of the subcomponents guarantees that, after measuring the performance of experiments on one subsystem, changes on this subsystem do not necessarily imply the expiration of the results on a higher level of component aggregation but result in a well-framed set of additional experiments.

On the other hand, the system and component's outcome may be evaluated semantically-wisely, with respect to the amount %AL + of?
of knowledge they contribute to increase awareness. %according to different functional levels, depending on the outcome semantics and on the outcome contribution to user awareness. 
In this case, we can distinguish experiments performed at data processing level (which do not necessarily require domain knowledge), at indicator level (where performance evaluation requires some domain contextualisation), and at scenario level (where the user is involved in the experiment, which unfolds according to a domain-driven plot).  

These two dimensions are complementary and may be leveraged to reduce the complexity of system evaluation, because performance evaluation is first done independently piece-wise along the two dimensions, then aggregated along the same dimensions to obtain the overall evaluation.

\begin{figure*}[th]
\begin{center}
  \includegraphics[width=12cm]{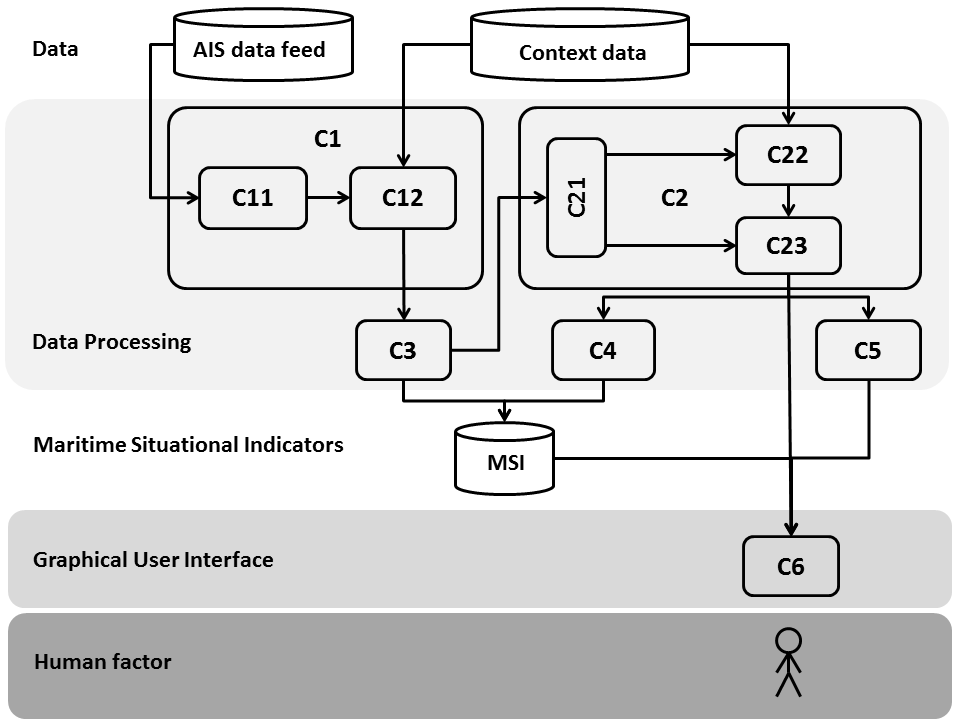}
\end{center}
\caption{Example of system architecture for a Maritime Situational Awareness System %\ec{it would be better to highlight the Big Data aspects} 
\label{fig:SystemArchitecture}}
\end{figure*}

An example of two-dimensional decomposition is given in Figure~\ref{fig:SystemArchitecture}, which schematically illustrates a Maritime Surveillance System. The software components implement the Data Processing and the Graphical User Interface layers of the architecture. 
At the interface layer, the user may be involved and domain-driven scenarios may be evaluated. 

At the intermediate layer, Maritime Situational Indicators (MSI) are examples of contextualised indicators that are outcome by subsystems and act as a semantic bridge between the software components and the domain. In Maritime Situation Awareness, MSIs are indicative of situations the operator or the analyst should be aware of to assess the %AL - scenario
situation. They may represent behavioural vessel events (high speed, gaps in communication, change of direction), whose collective evaluation is used by operators to assess potential maritime security threats. 

The system, which is designed to increase awareness, has different modules ($C_1$, ..., $C_6$) implementing different functionalities that altogether concur to the detection of MSI
%AL Maritime Situational Indicators (MSI) 
of interest. %for increasing awareness.
The data fed into the system varies in type and veracity. 

%----------------------------------------------------------------------------------------------------------

Note that DoEs differ for deterministic and non-deterministic sub-systems. In the following, the computational processes are assumed to be deterministic, if not stated differently. Only the subsystems including human factors are treated as non-deterministic. When deterministic and non-deterministic sub-systems are combined, the resulting system is treated as non-deterministic system with respect to the selection of DoE methods, as formalised by the following axiom:
\\
\\
$NonDeterministicSystem \equiv \exists hasComponent.NonDeterministic.$
\\

For the kind of big data systems discussed herein, we will discuss in the following semantic-wise decompositions for scenario level and data processing level evaluation. 
Scenario level evaluation, which requires decomposition involving the human factor, may be further differentiated if the evaluation assesses the graphical user interface or the entire big data solution. At data processing level, no human factor needs to be included in the decomposition, and in the case of a MSA system, data processing components may be evaluated using a different decomposition depending %AL - on 
if the components produce MSI %results 
or not.   

% For the kind of big data %datAcron 
% system discussed herein the following first two, semantic-wise,  decompositions will described in more detail:  %detailed 
% %subsequently:
% \begin{itemize}
% \item Decomposition of big data solution including human factor:
% \begin{itemize}
% \item Scenario level A: Human factor and graphical user interface
% \item Scenario level B: Human factor and entire big data solution
% \end{itemize}
% \item Decomposition of big data solution without human factor:
% \begin{itemize}
% \item Data processing components with MSI results %(SG, CER, CEF)
% \item Other data processing components  %(LED, SG, SI, DM, In-situ processing(LED,SG), (In-situ processing, SI),...)
% \end{itemize}
% \end{itemize}

Depending on the level of decomposition, different assumptions are necessary for stating and investigating possible research questions. Depending on the research question, the factors and factor levels are selected. 

\subsubsection{Scenario level: Human factor and entire big data solution}
The goal of the scenario level assessment is the quantification of the true and false detections of the expert user. 
The system boundary for the scenario level assessment includes the full system prototype and an expert user.

All components invisible to the expert user can be discarded from the scenario level assessment and treated separately in the MSI- or data analytics assessment in a %primary 
preliminary evaluation. A second evaluation on scenario level is necessary to evaluate the impact of the system latency on the degree of task fulfillment of the operator which needs to include all components.
For these reasons, the scenario level assessment focuses on the variation of MSIs, synopses and context information. Both variety %variation 
%AL - SYNOPSES have not been introduced yet
and veracity variations are considered. Volume and velocity variations are not considered explicitly, because in the considered application scenario, the ratio of vessels per expert user is guaranteed to be kept constant by the size of the control area.
%, here the STM area around Brest is selected as a reference. 
The fluctuations of the number of vessels inside the test data set is assumed to be representative for the test area.

In this setup, the case of a good performing expert user compensating a bad performance of the %datAcron 
system prototype can not be distinguished from the case of a bad performing expert user interacting with a good performing %datAcron 
prototype. These alternative interpretations \cite{Soldatova.2006} need to be addressed by the DoE \cite{Cook.1979} for strengthening internal validity %(e.g. is the difference between a collision and a non-collision significant?) 
and external validity.%(is the results of the experiment generalizable to non-experimental conditions?)
Thus, the data for the experiments is either taken from actual recordings or is modified in a minimalistic fashion and with respecting the coherence of data subsets, e.g. ocean conditions and vessel behavior. A detailed description of the method for the creation of data with external validity is subject to an upcoming publication.

%scale development process  \cite{Morgado.2017}:
%\begin{enumerate}
%\item item generation
%\item theoretical analysis/item refinement
%\item psychometric analysis: for construct validity, use EFA or CFA. 
%\end{enumerate} 
%As the number of participants is limited and possible confounding with unknown variables can not be excluded beforehand (e.g. reasons for knowing or not-knowing the daylight sign for 'movement ability affected') a collection of additional factors allows for alternative interpretations of the results (e.g. last time at sea).

% \item Scenario level A: Human factor and graphical user interface
\subsubsection{Scenario level: Human factor and graphical user interface}
For the exemplar characterisation of this subsystem, % 'Human factor and graphical user interface', 
the MSIs are considered as input. The research questions for this scenario level decomposition investigate the contribution of MSIs to the scenario specific task fulfillment and are presented in the following, bottom-up. 
%and can be divided bottom-up:%(e.g. to prompt zooming in)? 
For each research question, the reduction of the MSI combinations, corresponding to the high level of a 2-level DoE, is also described. 

After the identification of suitable MSI combinations, the underlying %AIS 
datasets and all eventual intermediate data aggregation states of upstream or downstream components %like synopses or low level events
are checked for conformance or exclusion respectively. All data subsets with conflicting interpretation, e.g. which relate to excluded MSIs, are removed from the experimental dataset.

%...Picture on the different data variations and performance measures assigned to the respective components%

% \begin{enumerate}
% \item Can the user assign a meaning to the symbols of the MSIs?
% \item Can the user distinguish situations in which the MSI fit the AIS data from situations in which the MSI do not fit?
% \item Can the user interpret a situation represented by multiple MSI?
% \item Can the user distinguish, thus prioritize situations or even assign different criticality estimates to different situations?
% \end{enumerate}

\paragraph{Research question: Can the user assign a meaning to the symbols of the MSIs?}
All MSIs need to be evaluated one by one, without combinations. %The MSI can be interpreted firstly on the paper without the offer of possible interpretations, then with interpretation. thirdly realtime in the visualization interface.

\paragraph{Research question: Can the user distinguish situations in which the MSI fit the AIS data from situations in which the MSI do not fit?}
%AL - Check the plural of MSIs?
Only a subset of MSIs needs to be evaluated. The necessary MSI combinations are selected by a domain expert who identifies the logical relationship between the MSIs: There are MSIs that are always (positively or negatively) correlated with each other. %Further, there are MSI that are likely to go (not) together in a specific situation (e.g. Under Way + On a maritime route). 
The selected subset of MSI needs builds the bases for the domain constraints derived from the following two research questions. For interchangeable MSI $A$ and $B$ the initial selection takes into account:\\
\\ 
$A \implies B$, \\ 
$A \implies \lnot B$, \\ 
$A \iff B$, \\ 
$\lnot A \iff \lnot B$ 

\paragraph{Research question: Can the user interpret a situation represented by multiple MSI?}
Only a subset of MSI needs to be evaluated. The selection of meaningful MSI combinations is effected according to instantiations of the respective scenario by a domain expert. As the generic scenario descriptions allow for the description of mutually excluding sub scenarios, the instantiation of those sub scenarios yield representative MSI combinations which simultaneously fulfill additional domain constraints impacting on the selection of DoE, e.g. the number of MSIs per vessel is unlikely to exceed three. The maritime domain specific constraints identified in 2) need to remain respected. 

\paragraph{Research question: Can the user distinguish, thus prioritise situations or even assign different criticality estimates to different situations?}
Only a subset of MSI needs to be evaluated. The prioritisation needs to be applied only on those situation descriptions which are interpreted correctly in the prior step.

\subsubsection{Data processing level}
There are two main goals for the evaluation on the data processing level. For the components that output MSIs, the main goal is the quantification of the veracity of the component results with respect to the combinations of true and false, positive and negative detections for each MSI.
The main goal of the evaluation of components located upstream from the MSI components is the quantification of the variation of the velocity and the volume of the data.

\subsection{Crossing input factors of components and data variations}
As all components are addressed by all dimensions of big data variations, a reduction of the complexity can be reached by the more detailed identification of infeasible combinations of input factor levels for each component. If the infeasible combination involves a connected component situated upstream or downstream, the corresponding factor levels on all connected components can be excluded. 
%it is supposed to be excluded the corresponding factor levels on all connected components. 
The constraint can be formalised in the ontology as a complex role inclusion axiom:\\
\\
$excludeEachOtherDerived \sqsubseteq isDerivedFrom \circ excludeEachOther \circ isDerivedFrom^{-}.$
\\
\\
An example for this restriction, is the exclusion of AIS data prompted by contradicting MSIs. %AIS 
Data which lead to contradicting MSI can be excluded from the design space. The exploitation of this input factor reduction is continued in the \ref{sec:OE}. Given the attainment of experimental results on the component level, the merging of those results to the system level are described in the following. 

\subsection{System recombination and result synthesis}
In the reverse sense of decomposition, the recombination of the performances of the subsystems to larger subsystems and finally to the entire system yield a reasonable performance hypothesis of the system. Further experiments are necessary to evaluate the interactions between recombined subsystems.

On the component level, each component that is evaluated can be assigned to a class with a specification of the effected experiment. Additional experiments on other component add knowledge on the same level. For deducing implication on larger subsystems i.a. on a higher level of recombination, the relationships between the evaluated components need to be taken into account. Following the description logic syntax, classes or concepts are identified by capitals while binary relationships or roles have a small first letter. For the deduction of subsystems consisting of both \textbf{interconnected and evaluated components}, a combination of local reflexivity and role compositions is proposed:
\\
\\
$EvaluatedComponent \equiv \exists evaluatedComponentRole.\textit{Self}. $	
\\
$evaluatedComponentIsConnectedTo \sqsubseteq evaluatedComponentRole \circ isConnectedTo \circ evaluatedComponentRole.   $
\\
$EvaluatedConnectedComponent \equiv \exists evaluatedComponentIsConnectedTo.\top.$
\\

The concept allows for an automated assertion of the components. Based on this pattern, sub concepts and sub properties can be defined for the refinement of the respective qualitative evaluation level reached, e.g. for formalising the binary status of requirement achievement. An extension for the quantification of the evaluation is introduced by the specification of the concept definition of $EvaluatedComponent$ to \textbf{$QuantitativelyEvaluatedComponent$}:
\\
\\
$QuantitativelyEvaluatedComponent \equiv EvaluatedComponent \sqcap \exists \ performanceMeasureGreaterThan.> "value". $	
\\

\begin{figure}[t]\vspace*{4pt}
%\centerline{\includegraphics{EquivalentClassAssertions}\hspace*{5mm}\includegraphics{EquivalentClassAssertions}}
\centerline{\includegraphics{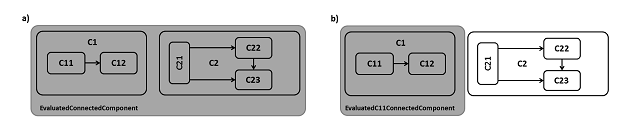}}
\caption{(a) Classification of all connected and evaluated components; (b) Classification of interconnected and evaluated components.\label{fig:EquivClassAssertion}}
\end{figure}

As shown in Fig. \ref{fig:EquivClassAssertion} a), this description does not differ between interconnected and separated but connected subsystems. Thus, for addressing only interconnected and evaluated components an additional discriminator is needed whose result is depicted in Fig. \ref{fig:EquivClassAssertion} b). For this, the definition of a connected component needs to be specified with respect to a reference component, e.g. C11: 
\\
\\
$EvaluatedC11ConnectedComponent \equiv EvaluatedConnectedComponent \sqcap isConnectedTo.C11.$
\\

Additionally, the role $isConectedTo$ has to be defined as transitive role so that components, both directly and indirectly connected to the reference component, are considered to fulfill the preceding definition. 
\\
Based on these axioms, different logic or arithmetic aggregation functions can be used to project the evaluation result from the component level to the next level of aggregation. Examples for aggregation functions for performance measures are completeness and throughput. Completeness is only given, if all components evaluated are complete. The throughput is defined by the component with smallest throughput of the system:
\\
\\
$Completeness_{EvaluatedConnectedComponent} = \bigcap \limits_{i=1}^n {C_i}.$
\\
$Throughput_{EvaluatedConnectedComponent} = Min(Throughput_{C_{i}}|C_{i} \in System).$
%~~~~~~~~~~~~~~~~~~~~~~~~~~~~~~~~~~~~~~~~~~~~~~~~~~~~~~~~~~~~~~~~~~~~~~~~~~~~~~~~~~~~~~~~~%
\section{An Evaluation Ontology for MSA information Sources and Experiments}\label{sec:OE}
%Formalising scientific and engineering experiments makes them reproducible, easier to share and search. It lets their results get connected to experimental outputs of future studies and accelerates their interpretation. In order to make these principles easily applicable, different possibilities for their formalisation are subject to ongoing efforts. The contribution of this article is the initialisation of an enrichment of priorly proposed concept descriptions, up to the point where a DoE can be proposed by a non-expert. The approach is not claiming comprehensiveness but advocates a stronger use of semantic web technologies for making DoE specific knowledge accessible to other domain descriptions.

In this section we propose a formalisation of domain specific experiments. 
%\subsection{Formalisation of domain specific experiments}
Starting from existing formalisations of experiments like EXPO, a subset of the description of scientific experiments consist of \cite{Soldatova.2006}:
\\
\dirtree{%This comment is required.
.1 ScientificExperiment. 
%.2 ExperimentalGoal.
.2 ExperimentalDesign.
.3 ExperimentalModel.
.4 TargetVariable.
.4 Factor.
.3 SubjectOfExperiment.
%.3 PlanExperimentalActions.
%.2 ExperimentalHypothesis.
.2 ExperimentalResults.
}
\
%This might help as well.yes
\\
In the following, the previously mentioned concepts are specified to the extend that examples of contradicting design decisions lead to inconsistencies in the ontology. This shall ease their detection and their replacement by consistent or even favorable design decisions.
\\
Starting from the first system decomposition in \ref{sec:SD}, two subjects of experiment can be distinguished, with characteristic properties: 1) {\em systems with human factor}, which are non-deterministic,  where a limited number of attributes should be considered, and where a controllable nuisance factor is given by the experience of experimentee; 2) {\em systems without human factor}, which are deterministic and no controllable or uncontrollable nuisance factors are included.

%with the following properties can be distinguished: 

% \begin{itemize}
% \item System with human factor
% \begin{itemize}
% \item non-deterministic
% \item human factor limitation on the number of attributes
% \item controllable nuisance factor: experience of experimentee
% \item uncontrollable nuisance factors: unknown
% \end{itemize}
% \item System without human factor
% \begin{itemize}
% \item deterministic
% \item no controllable nuisance factors
% \item no uncontrollable nuisance factors
% \end{itemize}
% \end{itemize}

For these two subjects of experiment the DoEs to choose are very different. As DoE is typically applied on non-deterministic processes, the fundamental principles of DoE are applicable to the system with human factor:\\
\\
$DoEWithReplication \equiv DoE \sqcap \exists hasSubjectOfExperiment.NonDeterministic.$\\
$DoEWithBlocking \equiv DoE \sqcap \exists hasNuisanceFactor.Controllable.$\\
$DoEWithRandomization \equiv DoE \sqcap \exists hasNuisanceFactor.Uncontrollable.$\\
\\
Given a controllable nuisance factor like 'years of experience at sea' or 'years of experience as maritime operator' the necessary use of blocking influences the number of experiments per experimentee. Additional restrictions including a maximum number of experiments per experimentee allows for a reduction of the candidate designs.\\
\\
$ DoEsuitableForHumanFactor \equiv \forall DoE.hasNumberOfExperiments.<"maximumNumberOfExperiments".$\\
\\
Tables including the design and the number of experiments can be used to instantiate the set of candidate designs, see e.g. \cite{cavazzuti.2012}. Designs which are using sampling methods where the experimenter can choose the number of experiments need to be distinguished differently. With an increasing number of restrictions, especially on the allowed combinations of factor levels, Optimal Designs are offering locally optimal Designs, given an optimality criterion.\\ 
\\
$OptimalDesigns \sqsubseteq DoEsuitableForFactorLevelConstraints $ \\
\\
D-Optimal Design is probably the most widespread Optimal Design, maximising the statistical efficiency. For the application in human factor experiments, the additional consideration of a maximum number of factors or attributes shown to the experimentee are an important criterion for maximising the consistency of the human factor responses, thus the external validity of the experiments \cite{Louviere.2008}. \\
\\
The DoE to choose depends on the effects to be tested, thus on the model of the response variable \cite{cavazzuti.2012}. If only main effects are of interest, e.g. a full factorial design may be used but normally implies with $NumberOfExperiments = FactorLevels^{NumberOfFactors}$ a much larger number of experiments than e.g. a Plackett-Burman design. Vice versa, a Plackett-Burman design does not allow for the separat estimation of main effects and all interactions, i.e. some effects can be confounded. For an experiment where the effect of interactions is not negligible, the use of such designs is not appropriate, thus excluded by the following description:\\
\\
Subsumptions: \\
$ModelWithInteractionEffects \sqsubseteq ExperimentalModel.$ \\
$ModelWithMainEffectsOnly \sqsubseteq ExperimentalModel.$\\
$OnlyMainEffectsWithoutConfounding \sqsubseteq ExperimentalModelOnlyMainEffects \sqcap NoConfounding.$\\
$PlackettBurmannDesign \sqsubseteq \forall isSuitableToModel.OnlyMainEffectsWithConfounding.$\\
\\
Covering axiom:\\
$ExperimentalModel \sqsubseteq ModelWithInteractionEffects \sqcup ModelWithMainEffectsOnly.$\\
\\
Disjointness of model concepts:\\
$ModelWithInteractionEffects \sqcap ModelWithMainEffectsOnly \sqsubseteq \bot.$\\
$ConfoundingDesign \sqcap NoConfoundingDesign \sqsubseteq \bot.$\\
\\
Accordingly, the use of a Plackett-Burman design excludes the selection of main effect models when confounding shall be excluded. Given the type of response variable, a similar procedure is applied to the model selection.

\section{Conclusions and Perspectives} \label{sec:concl}
The evaluation of big data systems %introduces 
%AL - envisages 
requires a large set of possible experiments due to the amount of different factors and factor levels. For the decomposition and recombination of large systems into subsystems, in this paper a unique procedure with different concepts is proposed  that allows  for the transfer of the subsystem results to the system level. Domain specific restrictions can be used to reduce the number of experiments in DoE by excluding infeasible or unnecessary experiments. As typically multiple domains are involved in the evaluation of a big data system, restrictions on multiple domains can be exploited in combination to restrain the set of relevant experiments. The proposed formalisation of restrictions may be used in order to %introduces the 
benefit of combining %these 
descriptions from different domains. Both for systems with human factors and computational processes the exclusion of inappropriate DoE is described axiomatically. 

The proposed domain restrictions can alternatively %AL + be
be learned from data. Thereby, the restrictions which refer to a perfect world can be extended and refined with restrictions from the actual, imperfect data (e.g. in the maritime domain, fishing vessels often use erroneously the AIS status {\em at anchor} while fishing, even though they are not anchored) and restrictions from the big data system (e.g. AIS status {\em at anchor} does not lead to MSI {\em at anchor or moored}, because the vessel is moving, but is additionally classified as {\em movement ability affected}, because trawling). Additionally, for the specification of fractional factorial designs, the mandatory selection of aliased effects of factors can benefit from the learning and formalisation of confounding rules from the domain specific data. Future work is also needed for describing the recombination of system properties whose values aggregate, e.g. latency or processing time. 

\section*{Acknowledgements}
This work is supported by the Big Data Analytics for Time Critical Mobility Forecasting (datAcron) project, which has received funding from the European Union’s Horizon 2020 research and innovation programme under Grant Agreement No. 687591.

%=REFERENCES=======================================================%
\small
\bibliographystyle{abbrv}
\bibliography{literature}

\end{document}